\documentclass[11pt,a4paper]{article}

\usepackage[margin=1in]{geometry}
\usepackage{graphicx}
\usepackage{booktabs}
\usepackage{amsmath,amssymb}
\usepackage{natbib}
\usepackage{hyperref}
\usepackage{xcolor}
\usepackage[expansion=false]{microtype}
\usepackage{enumitem}
\usepackage{xurl}
\graphicspath{{figures/}}
\hypersetup{colorlinks=true, linkcolor=blue!50!black, citecolor=blue!50!black, urlcolor=blue!60!black}

\title{How Should a Camera-First AI Agent Work?}

\author{Xiaofan Wu, Xi Zeng, Miaoxia Chen, Peishan Chen, Shuyan Li,\\
Chen Wang, Xiaohan Wang, Wanling Xu, Jiyun Yao, Hanyong Zhong, Jiahao Zhu\\
Chance AI\\
\texttt{xiaofan@chance.vision} (corresponding author)}

\date{}

\begin{document}
\maketitle

\begin{abstract}
Camera-first agents must act on the visual world as it unfolds. We argue that the central obstacle for future vision-interaction systems is not recognition ability---multimodal models (vision-language and live/streaming alike) generalize well through language and remain the right tool for semantic reasoning---but \textbf{time}. Many visual targets are \emph{ephemeral}: they appear, are query-relevant, and leave within a short window. Binding memory to the same end-to-end path as reasoning means that by the time a monolithic pipeline completes, the evidence is gone. We formalize this as \textbf{Ephemeral-Target Interaction} and a \textbf{Perceptual Deadline Contract} separating a hard \emph{memory} deadline from a soft \emph{reasoning} deadline, grounded in the bounded, passive nature of sensory and working memory. We propose a \textbf{two-deadline architecture}: an always-on, bounded, non-semantic fast path; an \textbf{active remembering} mechanism that commits a moment to durable memory before it is overwritten; and a \textbf{deferred deliberation} layer (multimodal model + tools). We instantiate the contract on real MOT17 track windows---a consistency check---and calibrate the single-path baseline from \textbf{measured} frontier-model latency: even a short answer takes $1.2$--$9.3$\,s, and an agentic identify-and-search longer. \textbf{Answers can be late; evidence cannot be missed.}
\end{abstract}


\section{Introduction}
Vision-language models have become remarkably capable because language provides a generalizing interface. For semantic recognition, retrieval, and reasoning, a strong VLM is currently the best tool we have, and we expect it to remain the core of the slow, deliberative layer of future vision systems. The difficulty of building a \emph{camera-first agent} lies elsewhere: \textbf{in time.}

Consider a camera-first agent interacting with the world in real time. Many of the targets a user cares about are \textbf{ephemeral}---a person passing by, a sign glanced at, an object briefly held. Such a target is present only within a short window. If perception and reasoning are bound into a single end-to-end path (see a frame, run a large model, identify, call tools, answer), then by the time the pipeline completes, the target has left and the evidence is gone. The user's tolerance is asymmetric: a late answer is tolerable; a missed moment is not. We summarize this as: \emph{answers can be late, evidence cannot be missed.}

This reframes what the next-generation intelligent terminal is. We take a \textbf{camera-first terminal} to be a device with \textbf{always-on but bounded perception}: it receives the visual stream passively and cheaply, while only \textbf{actively selected} moments are persisted. Its defining competence is not answering questions about an image but \textbf{not missing the moment}---a perception-first counterpart to today's prompt-first, stateless model-plus-screen devices. This is no longer hypothetical: VisionClaw \citep{liu2026visionclaw} already runs such an agent on Meta Ray-Ban glasses, yet samples perception at $\sim$1\,Hz and reports a median $\sim$12\,s action latency---exactly the temporal gap we address.

This asymmetry suggests that the fundamental design question for camera-first agents is not ``how do we recognize better?'' but \textbf{``what should the system do in the moment?''} We argue for a \textbf{Perceptual Deadline Contract} that separates two deadlines with different slack: a \textbf{hard memory deadline} (lay down the evidence within the target's window $\tau$) and a \textbf{soft reasoning deadline} (the semantic answer may arrive later, computed on frozen evidence).

We instantiate this as \textbf{a two-deadline architecture}: an always-on, non-semantic fast path that passively writes a bounded working memory; an \textbf{active remembering} mechanism that remembers a moment into durable memory before it is overwritten (event-driven in v1); and a \textbf{deferred deliberation} layer (a multimodal model with tools) that reads from memory. The multimodal model remains the slow, deliberative layer: we change \emph{where} and \emph{when} it runs, not whether it is used. The design is motivated by the bounded, automatic nature of sensory and working memory \citep{sperling1960brief,atkinson1968human,baddeley1974working,cowan2001magical} and by reactive-deliberative control \citep{brooks1986robust}. Perception is itself inference --- the visual system performs fast, automatic, non-linguistic inference \citep{rao1999predictive,kersten2004object} --- so the contract orders \emph{two kinds of inference by deadline}, not reasoning from non-reasoning.

\paragraph{Contributions.} (1)~\textbf{Problem:} we formalize Ephemeral-Target Interaction and the Perceptual Deadline Contract (hard memory vs.\ soft reasoning), grounded in bounded sensory/working memory and budget-conditioned remembering. (2)~\textbf{Formulation:} a deadline curve---memory recall versus dwell $\tau$---and an initial instantiation on real track windows. (3)~\textbf{System:} a two-deadline architecture (bounded passive store $+$ event-driven active remembering $+$ deferred deliberation), reported as memory recall vs.\ budget and a bounded-horizon sweep. (4)~\textbf{Benchmark:} an ETI benchmark on public MOT17 enter/exit tracks with an injected query per episode.

\paragraph{Non-goals.} No product lift, no CTR, no claim that multimodal models are the wrong substrate, no human-cognition isomorphism, and no user-experience claim in this version. Query/goal-conditioned selectivity and false-memory are deferred to a v2 selective benchmark.

\paragraph{Scope.} Our claim is scoped to current constraints---interaction latency, uplink bandwidth, and inference cost. We do not claim a single model is intrinsically insufficient: with near-free, instantaneous inference, a monolithic design would serve both stages and the contract would be unnecessary.

\section{Related Work and Positioning}
\textbf{Single-path VLM agents and streaming video models.} Recent camera-first and streaming systems route each frame or segment through one (large) multimodal model, optionally with tool calls. We cut on the \textbf{temporal contract}: which computation must complete \emph{within} a target's window, and which may be deferred.

\textbf{Realtime, omni, and full-duplex models.} A new class of streaming, low-latency multimodal models---GPT-4o Realtime, Gemini Live, Qwen2.5-Omni \citep{qwen25omni2025}, Moshi \citep{moshi2024}---provides always-on audio-visual streaming with simultaneous input/output, and is a natural substrate for both the fast perceiver and the deferred deliberator (VisionClaw already runs on Gemini Live). It remains, however, a \emph{single path} with an ephemeral context: continuous perception does not by itself guarantee that a sub-second target is remembered before it is lost, nor that evidence is retrievable for a later, tool-using answer. Our contract is \textbf{model-agnostic} and sits above this substrate: it fixes \emph{when} evidence must be remembered and \emph{what} must be persisted, however the backbone is built.

\textbf{Always-on wearable agents.} VisionClaw \citep{liu2026visionclaw} is the closest system: on Meta Ray-Ban glasses it streams egocentric video into Gemini Live and delegates actions to an agentic backend, with reported gains in task time and workload. Two of its own measurements frame our problem: perception is sampled at $\sim$1\,frame/s, and end-to-end latency has a median $\sim$12\,s. Under our contract, 1\,Hz sampling would leave sub-second windows largely uncovered, and 12\,s is far outside the memory window. We treat VisionClaw-style systems as the motivation and formalize the perceptual-timing layer they currently lack.

\textbf{Cascades, routing, and adaptive computation.} Fast/slow and conditional compute have deep precedent: classifier cascades \citep{viola2001rapid}, adaptive computation time \citep{graves2016adaptive}, early exit \citep{teerapittayanon2016earlyexit}, sparse MoE \citep{shazeer2017outrageously}, LLM cascades \citep{chen2023frugalgpt,aggarwal2023automix}, routing \citep{ong2024routellm}, per-token depth routing \citep{raposo2024mod}, and vision-internal reductions \citep{rao2021dynamicvit,bolya2023tome}. Our target differs: \textbf{event-level remembering under a hard memory deadline}.

\textbf{Dual-process and reactive-deliberative architectures.} Our fast/slow split echoes dual-process theories \citep{kahneman2011thinking,evans2013dual}; we use the ``System 1/2'' label only as a heuristic and make no cognitive-psychology claim \citep{melnikoff2018mythical}. Layered fast/slow control dates to subsumption \citep{brooks1986robust} and three-layer architectures \citep{gat1998on}. We revisit it for VLM-era camera-first agents.

\textbf{Attention, saliency, active vision.} Saliency \citep{itti1998saliency} and active perception \citep{bajcsy1988active,mnih2014ram} establish selective viewing, typically goal-agnostic; our remembering is event-driven in v1 and becomes query/goal-conditioned in v2.

\textbf{Bounded memory.} Iconic/sensory memory is brief and capacity-limited; remembering commits selected content to durable store \citep{sperling1960brief,atkinson1968human,baddeley1974working,cowan2001magical}. Used as design motivation, not isomorphism.

\textbf{Latency in HCI.} Response-time thresholds \citep{miller1968response,nielsen1993usability} motivate an interaction budget; our failure is stronger---lost evidence.

\textbf{Counter-evidence.} (i) \emph{Record everything}: under an unbounded buffer remembering is trivial; we condition every metric on a budget and treat record-everything as the oracle upper bound. (ii) \emph{Representation convergence} \citep{huh2024platonic}: semantic-level convergence does not make the perceptual fast path redundant. (iii) \emph{Fast VLMs}: our claim is about the temporal contract, not raw speed.

\section{Protocol and Metrics}
\paragraph{Task.} \textbf{Ephemeral-Target Interaction (ETI):} a target appears at $t_{\mathrm{in}}$, is query-relevant, and leaves at $t_{\mathrm{out}}$ with dwell $\tau=t_{\mathrm{out}}-t_{\mathrm{in}}$. The agent must (i) remember within $[t_{\mathrm{in}},t_{\mathrm{out}}]$ under memory budget $B$, and (ii) later answer the query from memory. Remembering is hard-deadline; the answer is soft-deadline.

\paragraph{Episode schema.} Each episode is $\{t_{\mathrm{in}}, t_{\mathrm{out}}, \text{target\_id}, \text{query}, \text{answer\_label}, \text{source}, \text{synthetic}\}$. Every episode carries an explicit query; without it, ``missed'' is undefined.

\paragraph{Budget.} All memory metrics are functions of a per-sequence memory budget $B\in(0,1]$: the fraction of ephemeral entries the gate may persist (a proxy for memory/compute). We keep $B$ (a capture quota) distinct from the passive buffer horizon $H$ (\S4.1, \S5.7): $B$ bounds how many entries may be persisted, $H$ bounds how long evidence remains available. The primary comparison is at fixed $B$; the oracle is unbudgeted and instantaneous.

\paragraph{Metrics (layered).} \emph{Problem:} Window Miss Rate (WMR) vs.\ $\tau$; latency. \emph{Solution (budget-conditioned):} memory recall; active-remembering rate; time-to-remember; bounded-horizon sweep. \emph{Guard:} deferred-answer accuracy $|$ remembered and joint success. False-memory/memory precision require a query with distractors and are deferred to v2.

\paragraph{Baselines.} monolithic VLM; always-fast; fixed-rate sampling; classic saliency gate; ours; oracle.

\section{Architecture: a two-deadline design}
\begin{figure}[t]\centering
\includegraphics[width=\linewidth]{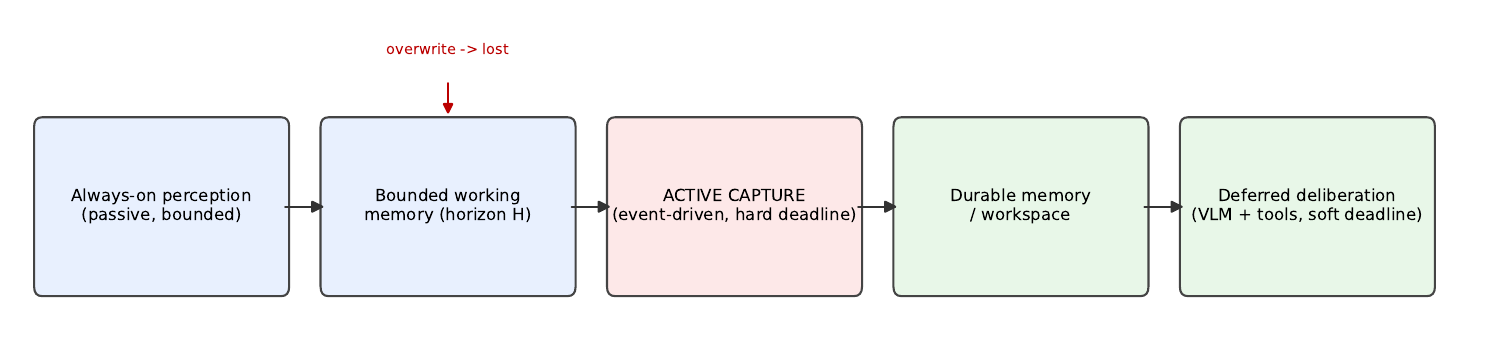}
\caption{Ours: always-on bounded passive store $\rightarrow$ active remembering (hard deadline) $\rightarrow$ deferred deliberation (soft deadline).}
\label{fig:arch}
\end{figure}

Our architecture instantiates the contract as three stages with a single live budget $B$ (Figure~\ref{fig:arch}).
\begin{itemize}[leftmargin=1.4em]
\item \textbf{Passive store (always-on, bounded, non-semantic).} Reception is passive: a cheap always-on path automatically writes a \emph{bounded} working memory with horizon $H$. It is not semantic and not deliberate; $H=\infty$ degenerates to the oracle. \emph{Always-on perception is not always-on recording.}
\item \textbf{Active remembering (event-driven, hard deadline).} A sub-second gate remembers a moment into durable memory before it is overwritten (v1: on target entry). Remembering must occur within $\tau$ and consumes $B$.
\item \textbf{Deferred deliberation (soft deadline).} A multimodal model with tools reads durable memory and answers; it may run after the target has left. Its output scores deferred-answer accuracy, not the primary claim.
\end{itemize}

\section{Experiments}
\paragraph{Benchmark.} Primary: \textbf{MOT17 enter/exit} (per-track hard windows; 7 sequences, 796 tracks; median dwell 3.67\,s; 369 tracks $\le 3$\,s) with an injected query. Secondary: LaSOT (re-entry), egocentric sources. Synthetic $\tau$-sweep overlay only as a controlled supplement.

\paragraph{Protocol.} For each episode, run a policy with budget $B$; log memory events and the deferred answer; report memory metrics as functions of $B$ and $\tau$. Deterministic; no fitting.

\paragraph{E1--E3 (design).} E1: deadline curve for the monolithic baseline. E2: ours vs.\ baselines (memory recall, TTR, compute; bounded-horizon sweep). E3: ablations (deferred). v1 ships E1 and the $B$- and horizon-sweeps as the labeled self-check (\S5.7); the rest is future work.

\subsection{Contract check on real windows}
\begin{figure}[t]\centering
\includegraphics[width=0.49\linewidth]{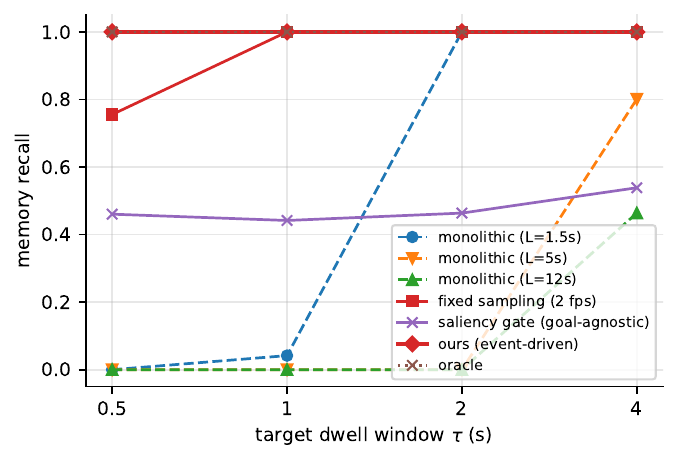}\hfill
\includegraphics[width=0.49\linewidth]{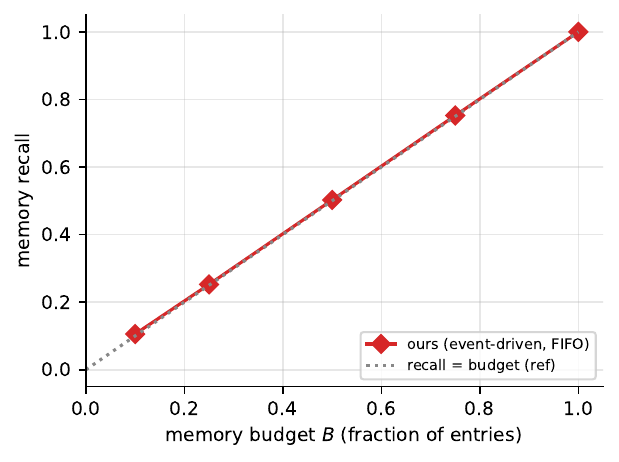}
\caption{Left: deadline curve (memory recall vs dwell $\tau$): the monolithic family collapses below its latency; ours/oracle are flat at 1. Right: memory recall vs memory budget $B$.}
\label{fig:deadline}
\end{figure}

\textbf{Perception is treated as free here.} Per-track windows are real (MOT17; 796 tracks, 369 $\le 3$\,s are the ephemeral core); with perception free, the check isolates the contract's \emph{scheduling} logic. TTR is a configured gate latency. Table~\ref{tab:main} and Figure~\ref{fig:deadline}.

\begin{table}[t]\centering\small
\caption{Memory recall on 796 real MOT17 tracks (scheduler simulation). Ours is event-driven at $B{=}1$.}
\label{tab:main}
\begin{tabular}{lcc cccc}
\toprule
policy & memory recall & TTR (s, conf.) & \multicolumn{4}{c}{recall @ $\tau=0.5/1/2/4$\,s}\\
\midrule
monolithic ($L{=}1.5$\,s) & 0.681 & 1.5 & 0.000 & 0.042 & 1.000 & 1.000\\
monolithic ($L{=}5$\,s)   & 0.430 & 5.0 & 0.000 & 0.000 & 0.000 & 0.801\\
monolithic ($L{=}12$\,s)  & 0.249 & 12.0 & 0.000 & 0.000 & 0.000 & 0.464\\
fixed sampling (2\,fps)    & 0.957 & 0.30 & 0.755 & 1.000 & 1.000 & 1.000\\
saliency gate (goal-agnostic) & 0.500 & 0.05 & 0.460 & 0.442 & 0.464 & 0.539\\
\textbf{Ours (event-driven)} & \textbf{1.000} & \textbf{0.05} & 1.000 & 1.000 & 1.000 & 1.000\\
oracle & 1.000 & 0.00 & --- & --- & --- & ---\\
\bottomrule
\end{tabular}
\end{table}

\begin{figure}[t]\centering
\includegraphics[width=0.5\linewidth]{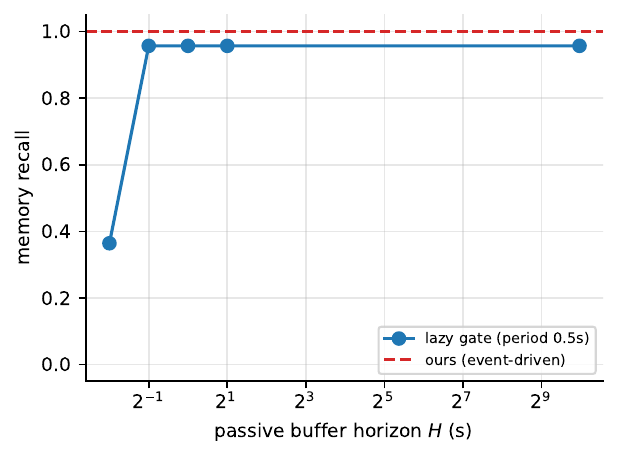}
\caption{Bounded-horizon sweep: memory recall vs passive buffer horizon $H$ (lazy gate, period 0.5\,s) vs the horizon-robust event-driven remembering.}
\label{fig:horizon}
\end{figure}

Reading. By construction, \emph{monolithic} recall is exactly the fraction of tracks longer than $L$ (the dwell survival function), event-driven remembering is $1.0$ (perception is free), goal-agnostic salience is a median split ($\sim$0.5), and the FIFO budget curve is $\mathrm{recall}=B$. The bounded-horizon sweep (lazy gate, period 0.5\,s) shows memory recall falling $0.957\to0.364$ as the horizon shrinks from 0.5\,s to 0.25\,s (Figure~\ref{fig:horizon}).

\subsection{Measured frontier-VLM latency (calibration)}
The useful latency of a monolithic pass is \textbf{estimated time-to-answer $=$ TTFB $+$ output\_tokens / throughput} ($+$ tool calls). TTFB alone is a floor. From OpenRouter's aggregated per-endpoint stats (last 30 min, request-weighted; text generation), estimated time-to-answer for a 64-token answer is 1.2--9.3\,s (Table~\ref{tab:lat}); low-throughput models pay most. This is a \textbf{lower bound}: OpenRouter's stats exclude \emph{image upload} and \emph{image-token prefill}; a frame is $\sim$100\,KB ($\sim$0.04--0.16\,s at 5--20\,Mbps) plus image tokens (hundreds to low-thousands, model-dependent). Both terms are positive, so the single-path failure is conservative. An agentic identify-and-search is longer still (VisionClaw: $\sim$12\,s median) \citep{liu2026visionclaw,openrouter2026latency}.

\begin{table}[t]\centering\small
\caption{Frontier-VLM latency (OpenRouter, last 30 min, request-weighted).}
\label{tab:lat}
\begin{tabular}{lccc}
\toprule
model & TTFB p50 (s) & throughput (tok/s) & est. time-to-answer, 64 tok (s)\\
\midrule
google/gemini-3.5-flash-lite & 0.73 & 127.5 & 1.23\\
openai/gpt-5.6-terra & 1.47 & 82.6 & 2.25\\
anthropic/claude-sonnet-5 & 1.63 & 70.3 & 2.54\\
openai/gpt-6-sol & 1.97 & 64.2 & 2.96\\
anthropic/claude-opus-5.5 & 5.50 & 65.7 & 6.47\\
qwen/qwen3-vl-8b-instruct & 3.92 & 12.0 & 9.25\\
\bottomrule
\end{tabular}
\end{table}

\section{Limitations}
\begin{itemize}[leftmargin=1.4em]
\item \textbf{Contingent on current costs.} The argument assumes a non-negligible latency/bandwidth/compute budget; with near-zero inference latency a single-path design would suffice---a judgment for the current environment, not a universal law.
\item \textbf{Oracle-perception simulation.} Policies run over real track windows but no pixels/detector/model are run; the scheduler is the object, not perception. The empirical section is a \emph{labeled self-check}, not a system evaluation.
\item \textbf{Latency is a floor.} Text-generation stats exclude image upload and image prefill; real vision latency is higher.
\item \textbf{Answer metric unrun.} MOT17 has no attribute labels; deferred-answer accuracy is not reported here.
\item \textbf{Query-conditioning/false-memory deferred} to a v2 selective benchmark.
\end{itemize}

\paragraph{Ethics and privacy.} Selective processing is selective data: the always-on fast path runs locally, and only actively remembered moments are persisted or uploaded, reducing exposure relative to continuous recording. This work uses public datasets only and no personal user data.

\section{Conclusion}
The binding constraint for future camera-first agents is not recognition quality but \textbf{time}. Many targets are ephemeral; a single end-to-end path loses their evidence before it can be used. We formalized this as Ephemeral-Target Interaction and a Perceptual Deadline Contract, and proposed a two-deadline architecture: a bounded passive working memory, an active remembering gate that remembers within the hard window, and a deferred VLM-and-tools layer that answers within the soft window. On real track windows, event-driven remembering recovers every ephemeral target while a monolithic pass misses all those shorter than its latency. The contribution is a measurable contract and a system that respects it: \emph{answers can be late; evidence cannot be missed.}

\bibliographystyle{plainnat}
\bibliography{refs}

\end{document}